\documentclass[final,5p,times,twocolumn]{elsarticle}

\usepackage{amssymb}
\usepackage{amsmath}

\usepackage{booktabs}
\usepackage{xcolor}
\usepackage{colortbl}
\usepackage{makecell}
\usepackage{multirow}
\usepackage{xspace}
\usepackage[colorlinks]{hyperref}
\newcommand{\ie}{i.e.,\xspace}

\journal{Pattern Recognition Letters}

\begin{document}

\begin{frontmatter}



\title{Advancing Video Self-Supervised Learning via Image Foundation Models}


%
\affiliation[a]{
    organization={School of Electronic, Electrical, and Communication Engineering, University of Chinese Academy of Sciences},
    addressline={Yuquan Road, 19A, Shijingshan District}, 
    city={Beijing},
    postcode={100049}, 
    state={Beijing},
    country={China}
}
\affiliation[b]{
    organization={StepFun},
    addressline={Suzhou Street, 3, Haidian District}, 
    city={Beijing},
    postcode={100080},
    state={Beijing},
    country={China}
}
\affiliation[c]{
    organization={Department of Automation, Tsinghua University},
    addressline={Shuangqing Road, 30, Haidian District}, 
    city={Beijing},
    postcode={100084}, 
    state={Beijing},
    country={China}
}

\author[a]{Jingwei Wu}
\ead{wujingwei22@mails.ucas.ac.cn}

\author[b]{Zhewei Huang}
\ead{hzwer@pku.edu.cn}

\author[c]{Chang Liu\corref{cor1}}
\ead{liuchang2022@tsinghua.edu.cn}
\cortext[cor1]{Corresponding author.}

\begin{abstract}
In the past decade, image foundation models (IFMs) have achieved unprecedented progress. 
However, the potential of directly using IFMs for video self-supervised representation learning has largely been overlooked. %
In this study, we propose an advancing video self-supervised learning (AdViSe) approach, aimed at significantly reducing the training overhead of video representation models using pre-trained IFMs. 
Specifically, we first introduce temporal modeling modules (ResNet3D) to IFMs, constructing a video representation model. 
We then employ a video self-supervised learning approach, playback rate perception, to train temporal modules while freezing the IFM components.
Experiments on UCF101 demonstrate that AdViSe achieves performance comparable to state-of-the-art methods while reducing training time by $3.4\times$ and GPU memory usage by $8.2\times$.
This study offers fresh insights into low-cost video self-supervised learning based on pre-trained IFMs.
Code is available at \href{github.com/JingwWu/advise-video-ssl}{\color{magenta}github.com/JingwWu/advise-video-ssl}.
\end{abstract}


\end{frontmatter}



\section{Introduction}
\label{sec:intro}

%
A foundation model refers to a pre-trained neural network, meticulously designed to serve a multitude of subsequent tasks. Image foundation models (IFMs)~\cite{radford2021learning, DBLP:conf/icml/RadfordKHRGASAM21}, which are trained on extensive and diverse datasets, have demonstrated remarkable proficiency in visual tasks.
%
%

Video self-supervised learning (SSL)~\cite{schiappa2023self} defines a label-free method that focuses on constructing video representations by using pretext tasks to aggregate spatio-temporal information.
While self-supervised spatio-temporal pre-training entails significant computational costs~\cite{feichtenhofer2021large, tong2022videomae}, IFM offers an alternative scheme for efficient learning of video representations by leveraging its spatial representation capabilities and shifting the focus to modeling temporal information. 

%
Previous studies explored distilling IFM's knowledge for temporal modeling~\cite{castro2022fitclip}, or incorporating IFM directly into self-supervised temporal tasks for joint training~\cite{dave2024no}.
While these approaches show promise, there remains ample room in fully harnessing IFM's spatial modeling capabilities and optimizing training efficiency.
Notably, in end-to-end supervised learning settings, IFMs have been successfully employed as visual backbones, directly utilizing spatial features and demonstrating their potential in video analysis~\cite{DBLP:conf/cvpr/ParkLS23, yang2023aim, ZhangZZSWZ24}.
These observations highlight the importance of rethinking the role of IFMs in video self-supervised learning in order to unlock their full potential for efficient and effective video representation learning.

\begin{figure}[!t]
    \centering
    \includegraphics[width=1.0\linewidth]{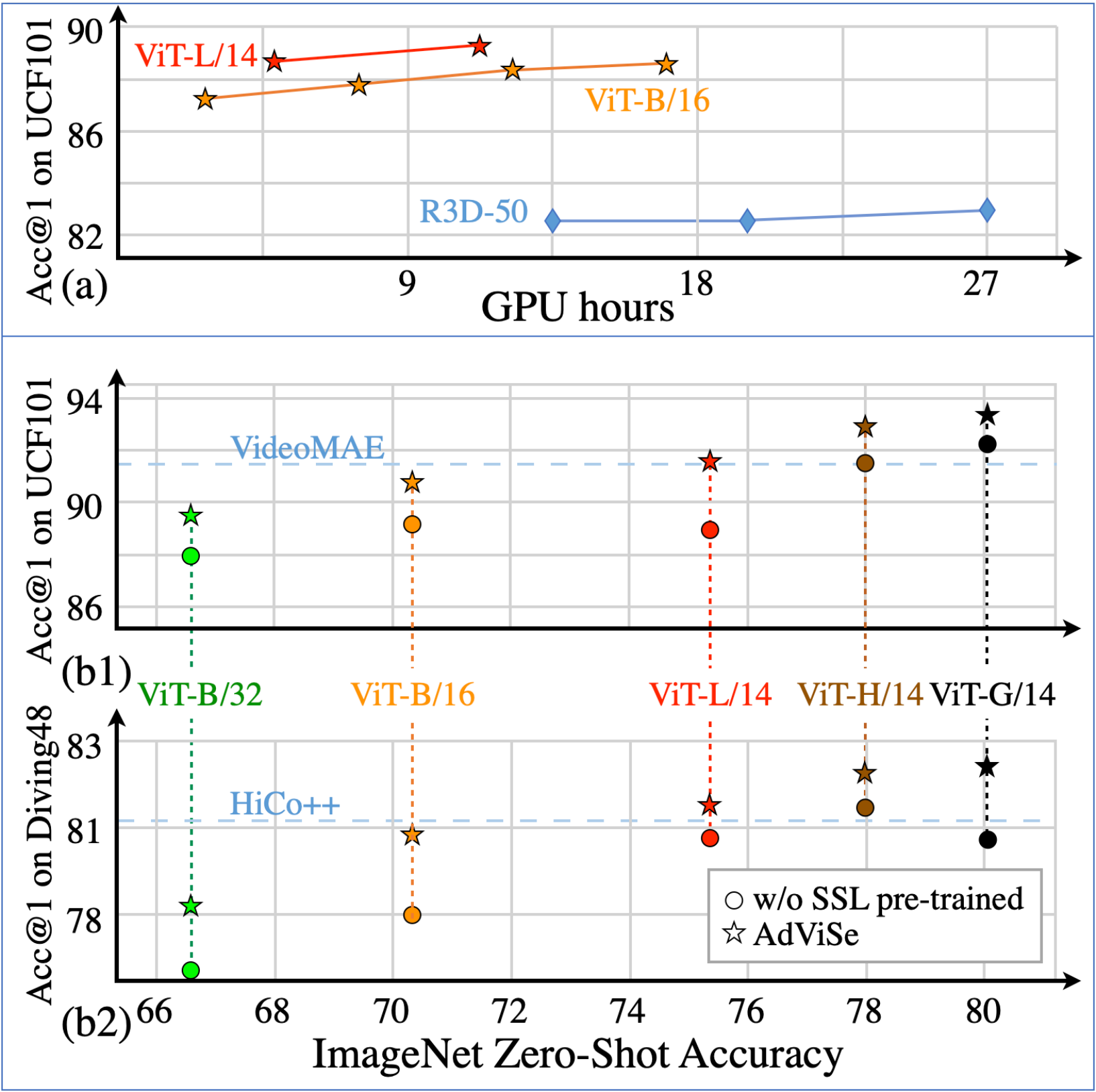}
    \caption{\textbf{AdViSe utilizes IFMs to implement efficient video self-supervised learning.} (a) With a video SSL method~\cite{yao2020video}, AdViSe significantly improves performance upon R3D-50~\cite{feichtenhofer2021large} with much lower training costs. (b1, b2) As IFMs~\cite{radford2021learning} enhance their spatial feature encoding capabilities (as indicated by ImageNet~\cite{deng2009imagenet} Zero-Shot Accuracy), the performance of downstream task (action recognition) also improves (AdVise). 
    }
    \label{fig:main-fig}
    \label{fig:train-eff}
\end{figure}

In this study, we reveal principled guidelines for spatial feature utilization ({SFU}) and temporal modeling module ({TMM}) design when incorporating IFMs into video self-supervised learning. We propose a new paradigm, referred to as {AdViSe}, where the parameters of IFM are frozen, and a lightweight TMM is pre-trained with an SSL pretext task to obtain temporal aggregation ability.
This approach integrates latent temporal information across multi-frame spatial features, constructing a comprehensive video representation efficiently.
We emphasize AdViSe's robust scalability, enabling performance improvements with enhanced IFMs, Fig.~\ref{fig:main-fig}.

We demonstrate that different video downstream tasks, whether focused on spatial or temporal understanding, can benefit from distinct configurations in both SFU and TMM design.
As for {SFU}, we focus on spatial resolution compression and channel dimension compression, while for {TMM}, we investigate network depth and module channel width. These studies lead to some design principles for a lightweight yet efficient IFM-TMM video SSL method.
%
AdViSe achieves performance comparable to previous video SSL methods while significantly reducing training costs. For example, experiments on UCF101~\cite{soomro2012ucf101} show that AdViSe achieves competitive performance with a $3.4\times$ reduction in pre-training time and a $8.2\times$ decrease in memory usage (detailed in Section~\ref{sec:effect}), highlighting both performance and efficiency gains of our method.
%

The contributions of this study are threefold: 
\begin{enumerate}
    \item We offer a comprehensive reassessment of spatial feature utilization and temporal modeling module, shedding light on the potential of integrating IFMs into video self-supervised learning.
    \item We introduce AdViSe for temporal modeling based on spatial features, achieves comparable performance with minimal tunable parameters and training costs.
    \item We validate the feasibility and efficiency of utilizing IFMs for spatio-temporal representation learning, establish performance baselines for future research.
\end{enumerate}

\section{Related Work}
\label{sec:related}

\paragraph{Image Foundation Model}
To adapt IFMs~\cite{radford2021learning, DBLP:conf/icml/RadfordKHRGASAM21} for video tasks, \textit{post-pre-training} techniques are employed on large video datasets~\cite{castro2022fitclip}. To mitigate the computational burden of retraining the entire network, these models can be frozen, with additional lightweight layers, known as \textit{adapters}, being updated during fine-tuning for temporal modeling. Similarly to adapters, \textit{prompt tuning} enhances efficiency by only fine-tuning a few additional parameters. These methods bridge the gap between images and videos through teacher-student distillation frameworks~\cite{castro2022fitclip}. However, the lack of video SSL upon IFMs hinders the exploitation of advancements in image encoder adaptation.

\begin{figure}[!t]
    \centering
    \includegraphics[width=0.95\linewidth]{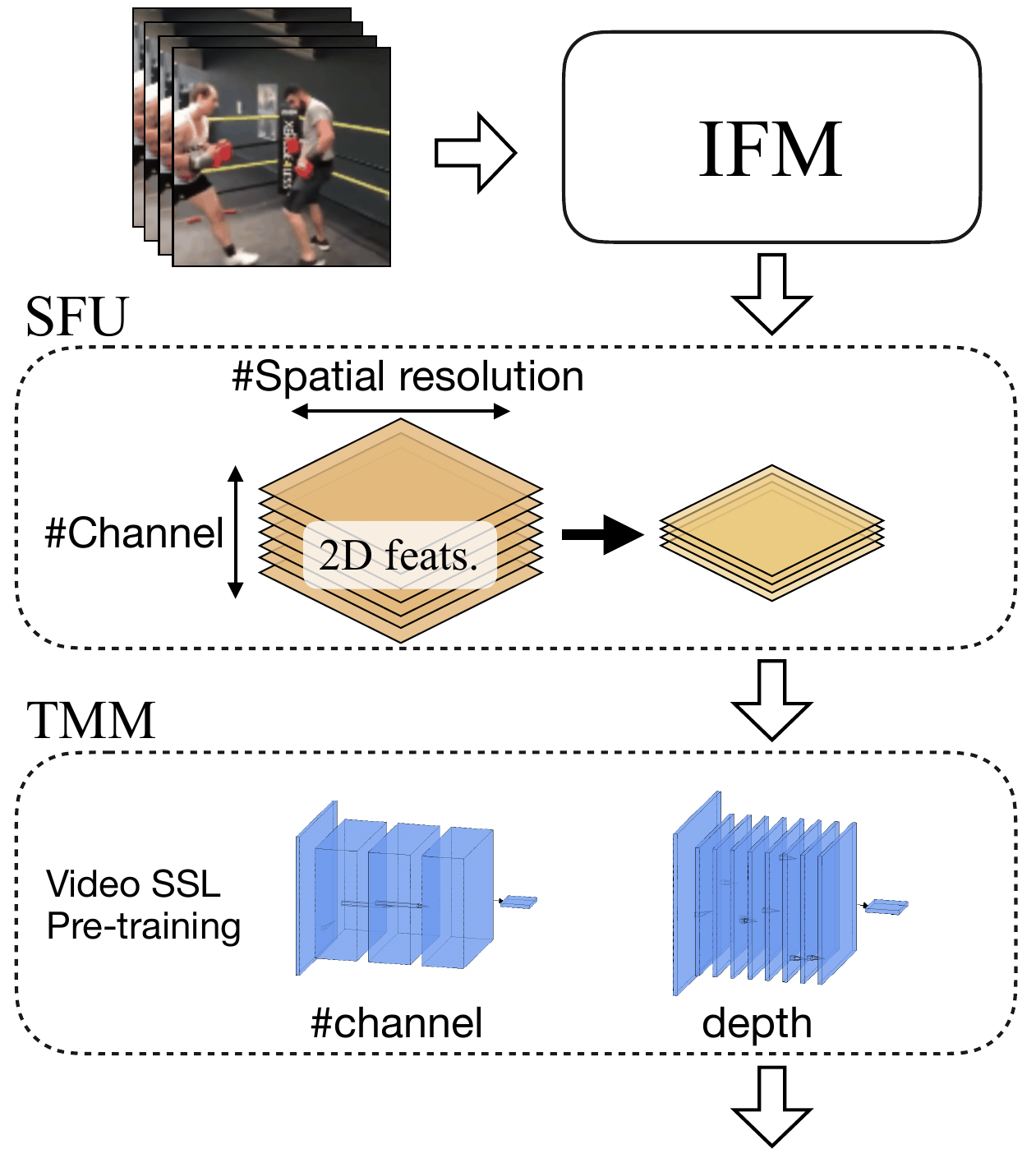}
    \caption{\textbf{AdViSe paradigm.} It leverages the spatial features produced by the image foundation model (IFM) to train the temporal modeling module (TMM), aggregating temporal information and effectively encoding motion dynamics.}
    \label{fig:pipeline}
\end{figure}

\paragraph{Video Self-Supervised Learning} 
This approach learns video representations by exploiting pretext tasks that create temporal and spatial coherence, such as predicting appearance statistics, classifying temporal order, and solving video jigsaw puzzles \cite{schiappa2023self}. 
In contrast, contrastive learning focuses on generating positive and negative samples to refine the distribution of spatio-temporal representations.
Popular image-based methods for generating positive and negative samples have been extended to the video domain by applying identical perturbations across video frames~\cite{he2020momentum, grill2020bootstrap, feichtenhofer2021large, chen2020simple}. 
Masked modeling involves randomly masking space-time patches and trains an auto-encoder using images and video streams during pre-training~\cite{tong2022videomae, wang2023videomae, wang2022bevt, ZhengLLLXWZC24}.
Despite the research progress, existing methods have largely overlooked the efficiency of video SSL, and optimizing computational efficiency remains critical for accelerating algorithmic advancements.

\paragraph{Architectural Design for Spatio-temporal Modeling}
The architectural evolution to incorporate spatio-temporal information begins with pioneering works~\cite{yue2015beyond, tran2015learning} operate in both spatial and temporal dimensions simultaneously. After that, I3D~\cite{carreira2017quo}, P3D~\cite{qiu2017learning}, R(2+1)D~\cite{tran2018closer}, and S3D~\cite{xie2018rethinking} extend 2D ConvNet~\cite{szegedy2015going} into 3D ConvNet, and achieve favorable speed-accuracy trade-off.
In parallel, transformers have gained prominence in video tasks, with architectures like VTN~\cite{neimark2021video}, TimeSformer~\cite{bertasius2021space}, ViViT~\cite{arnab2021vivit}, and Video Swin Transformer~\cite{liu2022video}, introducing temporal attention layers atop pre-trained Vision Transformers (ViTs), thereby offering a blueprint for temporal modeling using pre-trained IFMs. 
These developments collectively inform the design of temporal modeling techniques that leverage pre-trained spatial representations, inspiring the efficient integration of IFMs in Video SSL frameworks.





\begin{figure*}[t!]
    \begin{minipage}{0.49\textwidth}
        \centering
        \begin{minipage}{0.49\textwidth}
            \centering
            \includegraphics[width=1.0\linewidth]{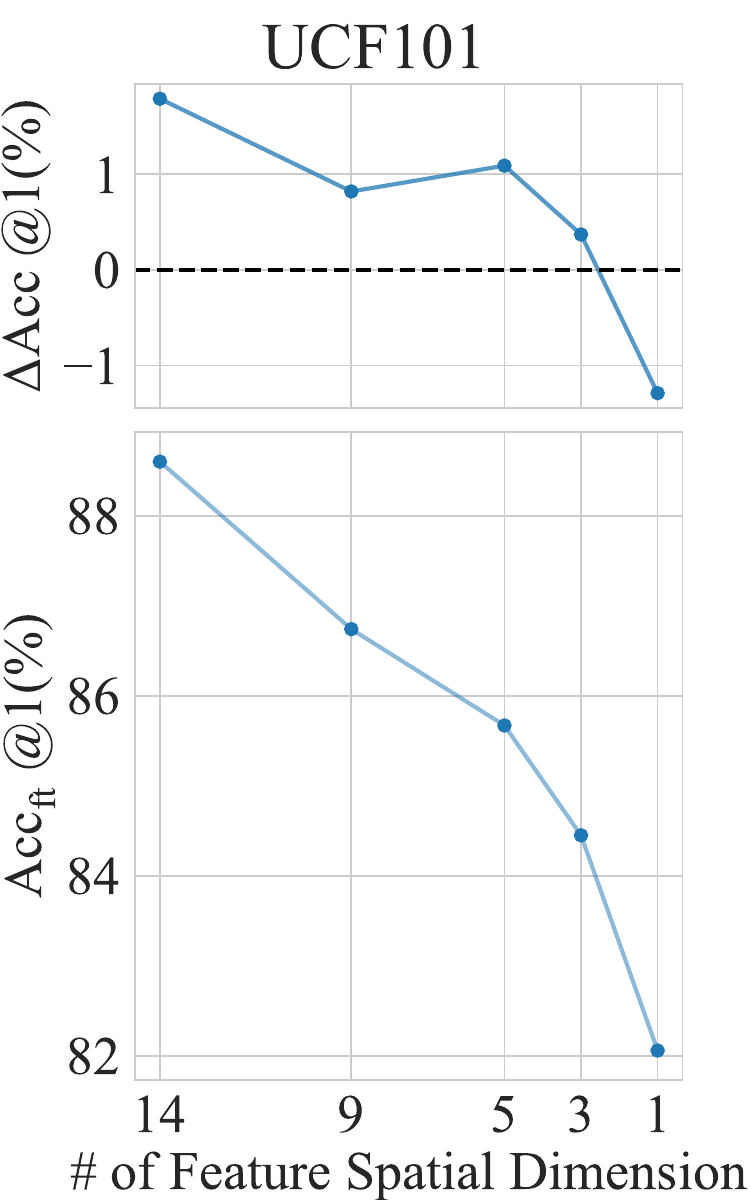}
        \end{minipage}
        \hspace{-2mm}
        \begin{minipage}{0.49\textwidth}
            \centering
            \includegraphics[width=1.0\linewidth]{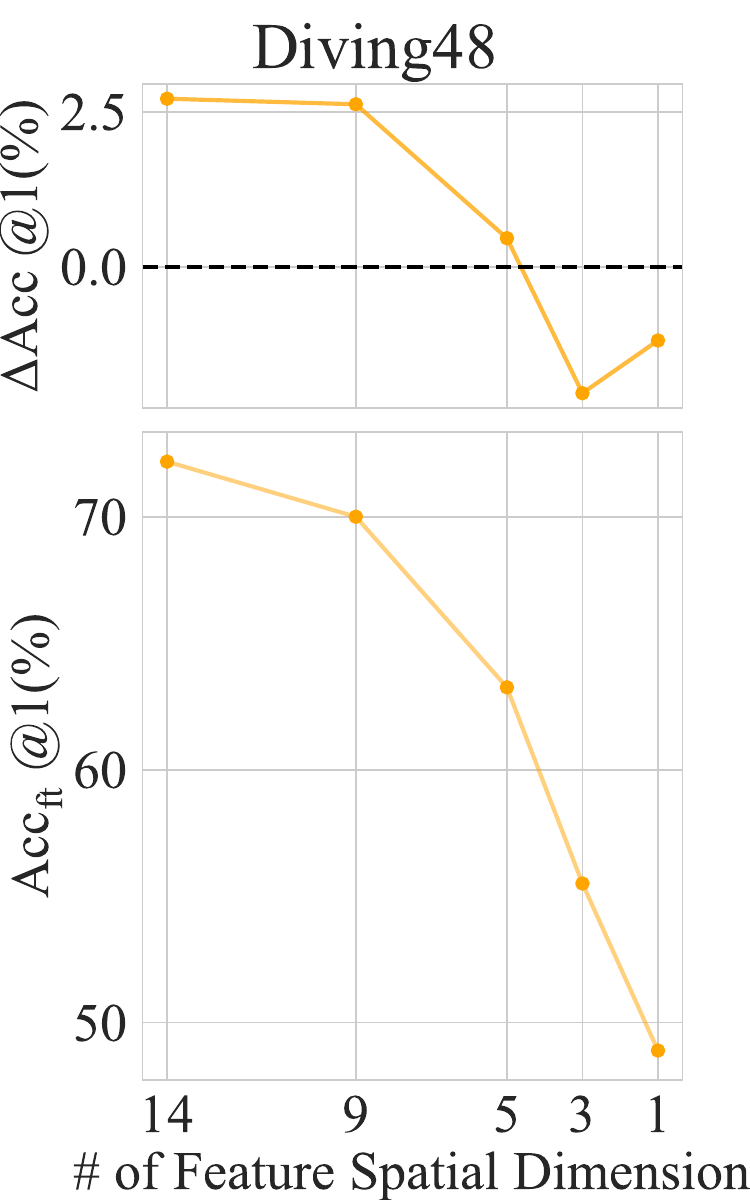}
        \end{minipage}
        \caption{\textbf{Impacts of Spatial resolution compression.} Spatial resolution compression significantly reduces the $Acc_{ft}$ metric (bottom) on Diving48~\cite{li2018resound}, but had a lesser impact on UCF101~\cite{soomro2012ucf101}, while the ${\Delta}Acc$ metric (top) shows a decline across both datasets.}
        \label{fig:spatial-compress}
    \end{minipage}
    \hfill
    \begin{minipage}{0.49\textwidth}
        \centering
        \begin{minipage}{0.49\textwidth}
            \centering
            \includegraphics[width=1.0\linewidth]{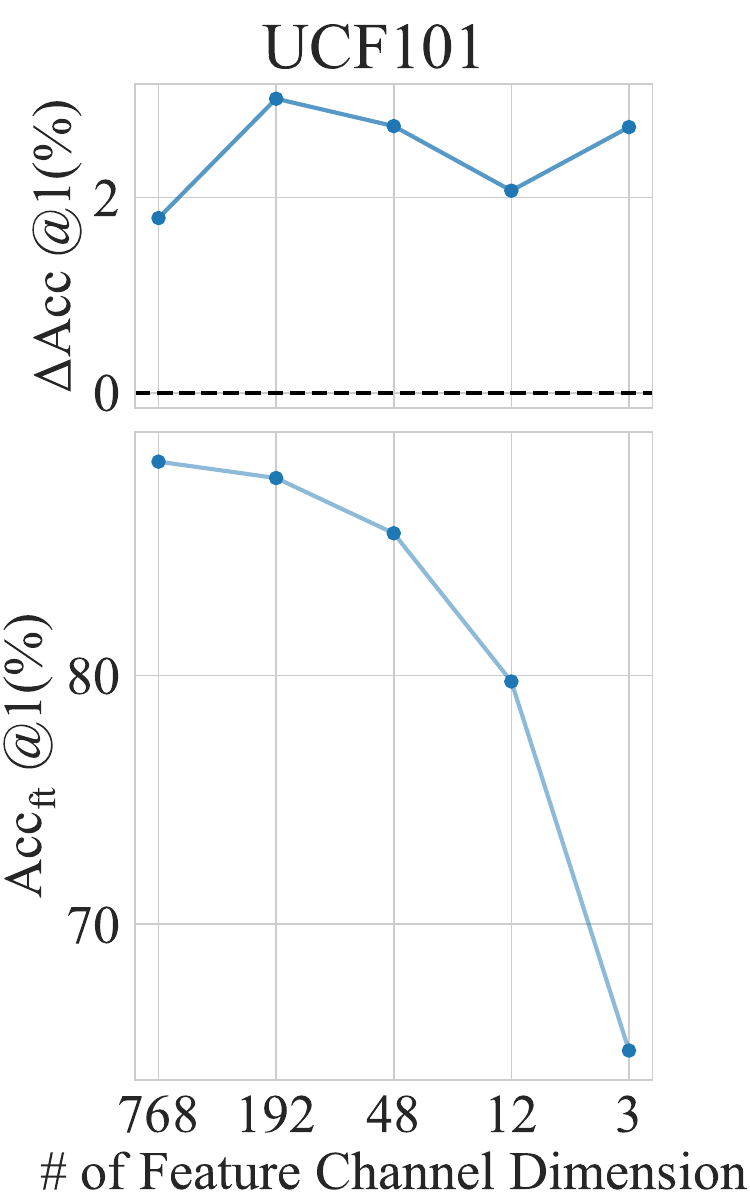}
        \end{minipage}
        \hspace{-2mm}
        \begin{minipage}{0.49\textwidth}
            \centering
            \includegraphics[width=1.0\linewidth]{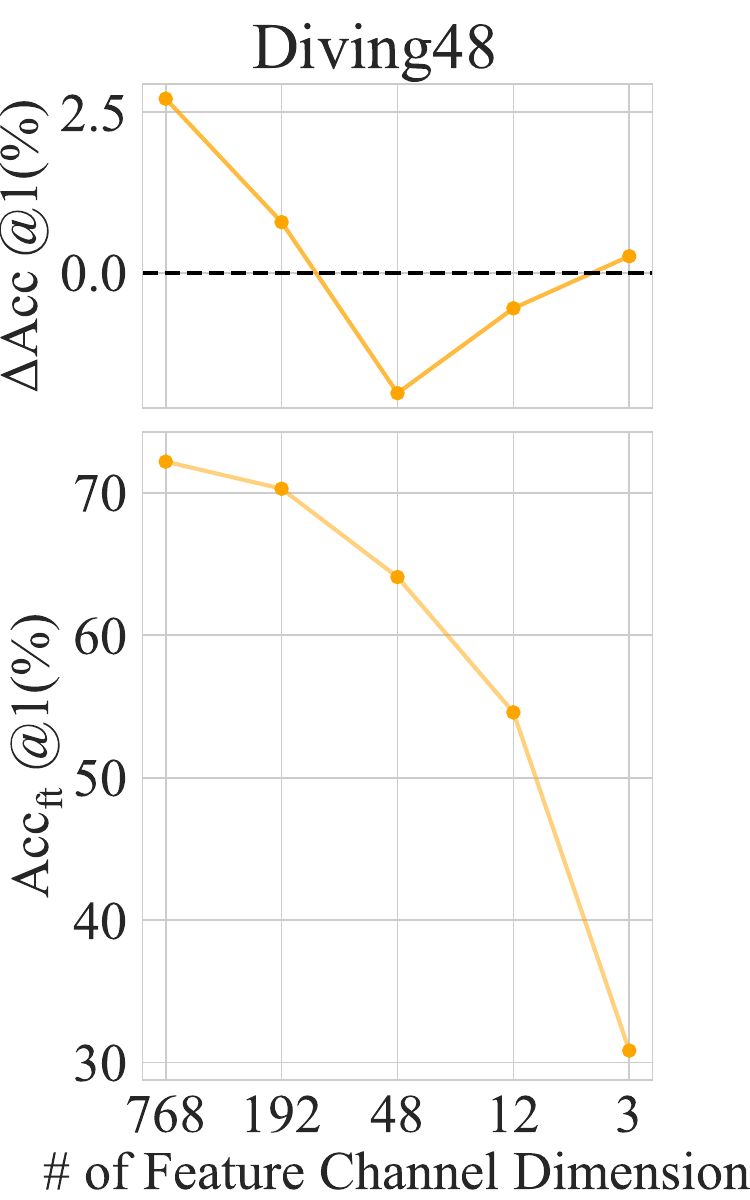}
        \end{minipage}
        \caption{\textbf{Impacts of Channel dimension compression.} Channel dimension compression reduces $Acc_{ft}$ performance (bottom) on both UCF101~\cite{soomro2012ucf101} and Diving48~\cite{li2018resound}. For ${\Delta}Acc$ metric (top), only Diving48 experiences some impact.}
        \label{fig:channel-compress}
    \end{minipage}
\end{figure*}

\begin{figure*}[htbp]
    \begin{minipage}{0.49\textwidth}
        \centering
        \begin{minipage}{0.49\textwidth}
            \centering
            \includegraphics[width=1.0\linewidth]{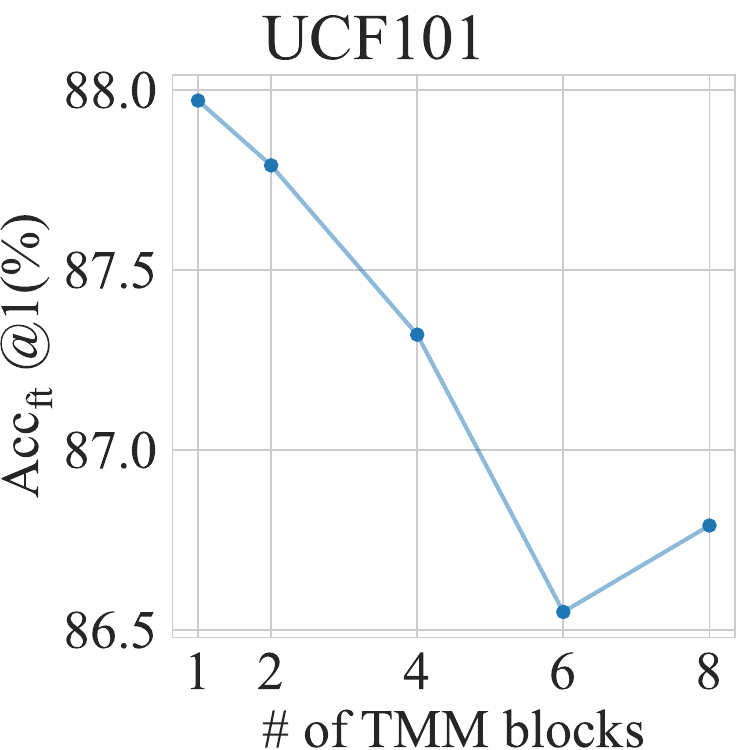}
        \end{minipage}
        \hspace{-2mm}
        \begin{minipage}{0.49\textwidth}
            \centering
            \includegraphics[width=1.0\linewidth]{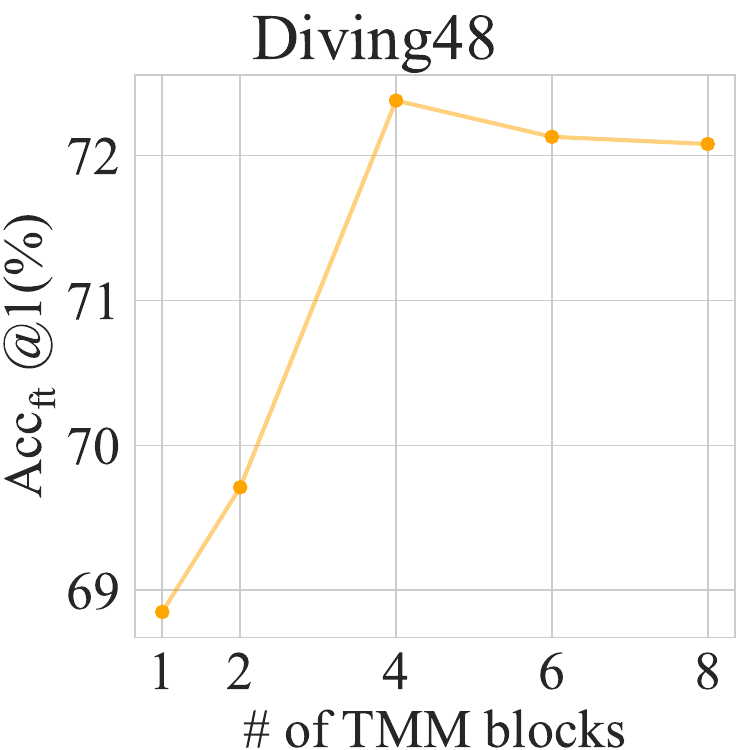}
        \end{minipage}
        \caption{\textbf{Impacts of TMM blocks.} Increasing TMM block number improves the accuracy of fine-tuning ($Acc_{ft}$) on Diving48~\cite{li2018resound} but reduces it on UCF101~\cite{soomro2012ucf101}.}
        \label{fig:incre-block}
    \end{minipage}
    \hfill
    \begin{minipage}{0.49\textwidth}
        \centering
        \begin{minipage}{0.49\textwidth}
            \centering
            \includegraphics[width=1.0\linewidth]{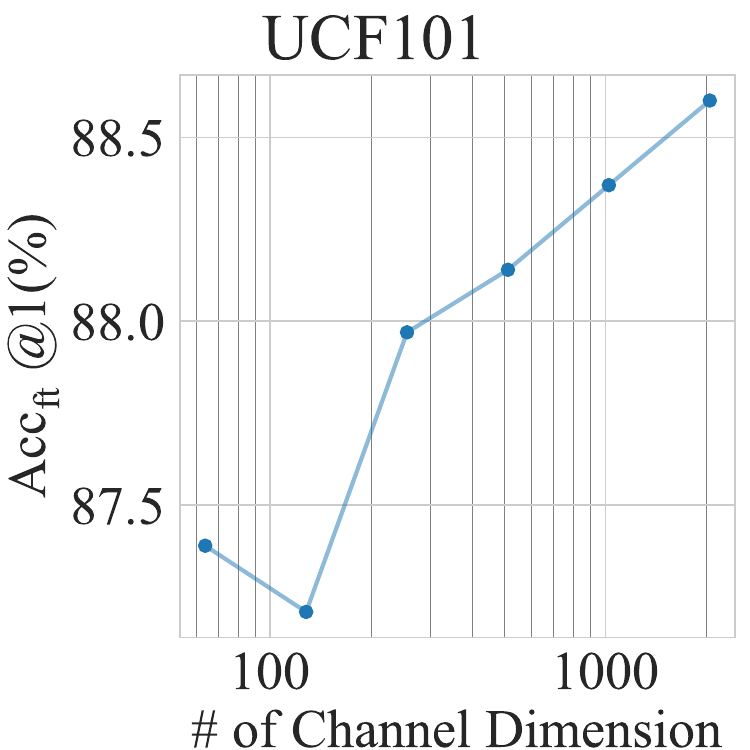}
        \end{minipage}
        \hspace{-2mm}
        \begin{minipage}{0.49\textwidth}
            \centering
            \includegraphics[width=1.0\linewidth]{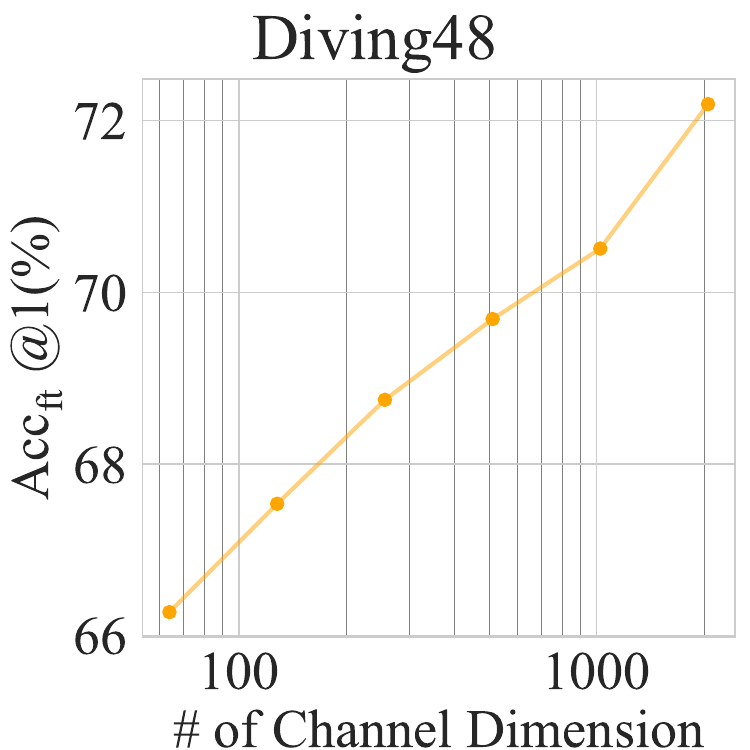}
        \end{minipage}
        \caption{\textbf{Impacts of TMM channels.} Increasing the hidden channel dimension of TMM enhances the $Acc_{ft}$ both on UCF101~\cite{soomro2012ucf101} and Diving48~\cite{li2018resound}.}
        \label{fig:incre-channel}
    \end{minipage}
\end{figure*}

\section{The Proposed Approach}
\label{sec:investigate}

\subsection{Model Architecture}

The AdViSe architecture consists of an image foundation model (IFM) and a temporal modeling module (TMM), Fig.~\ref{fig:pipeline}. Frame-wise representations from the IFM are processed by the SFU module before being fed to TMM. TMM is pre-trained through a self-supervised pretext task. During downstream fine-tuning, the parameters of TMM are inherited, and a task head is trained from scratch. The IFM remains frozen during both pre-training and downstream fine-tuning.
%
%

%

We utilize R3D blocks as TMM by default. A block comprises three layers of Conv3D with kernel sizes of $ 3 \times 1 \times 1$, $ 1 \times 3 \times 3$, and $ 1 \times 1 \times 1$, respectively. The number of hidden layer channels in R3D block is set to $256$. TMM does not down-sample the input IFM feature map in either the temporal or spatial dimensions, so the output feature map from TMM retains the resolution of IFM's outputs. An MLP (2 FCs with hidden dimension $1024$) head follows TMM to aggregate information across spatio-temporal dimensions in pre-training and fine-tuning phases.

\subsection{Model Training}

\paragraph{Pretext Task}
Playback Rate Perception (PRP)~\cite{yao2020video} excels in modeling time-domain information by training a model to distinguish input sample down-sampling rates, fostering sensitivity to temporal dynamics. We use it as a pretext task for video self-supervised learning.

\paragraph{Datasets}
We evaluate AdViSe across various video benchmarks, which are coarsely categorized into two.
(i)~Benchmarks that emphasize spatial information: {UCF101}~\cite{soomro2012ucf101} contains $101$ action categories, totaling $13,320$ videos. {HMDB51}~\cite{kuehne2011hmdb} consists of $6,766$ video clips spanning $51$ action categories. {Kinetics-400}~\cite{carreira2017quo} (K400) is a large-scale action recognition dataset comprising $240,553$ video clips, covering $400$ human action categories.
(ii)~Benchmarks that emphasize temporal information: {Diving48}~\cite{li2018resound} is a fine-grained video classification dataset, with around $18,000$ trimmed video clips covering $48$ different diving sequences. Diving48 is designed to minimize spatial bias~\cite{li2018resound}, making it necessary for the action recognition task to require greater temporal understanding. {Something-Something V2}~\cite{goyal2017something} (SSv2), with over $220,000$ videos across $174$ action categories, is designed for evaluating models in human action and gesture comprehension.


\paragraph{Pre-training}
We pre-train AdViSe using UCF101 for $800$ epochs ($200$ epochs when training on K400). During training, $4$ clips are randomly sampled from each video ($16$ clips sampled on K400), with each clip consisting of 8 frames at a resolution of $224\times224$. AdamW optimizer~\cite{loshchilov2017fixing} is employed with a $\beta$ value of $(0.9, 0.999)$. The learning rate coefficient $\eta$ is set to $10^{-2}$, and the actual maximum learning rate is calculated as $ \eta \times b / 64$ with cosine annealing to $0$. Weight decay is set to $10^{-6}$. Unless otherwise specified, we freeze the parameters of IFM and train  TMM and task heads. We use $32$ A800 GPUs for pre-training with a default batch size of $32$ videos per GPU.

\paragraph{Fine-tuning}
On UCF101, we randomly sample a $16$-frame clip from one video at $224\times224$ and fine-tune the TMM and classification head for $120$ epochs. We use the AdamW optimizer with a base learning rate of $10^{-2}$ and a weight decay of $5\times 10^{-3}$. The learning rate setting follows the pre-training stage. We follow the large-scale training practice~\cite{feichtenhofer2021large}  and report the $30$-view top-$1$ accuracy. On Diving48, the epoch number is set to $60$, and $10$ clips are uniformly sampled from each video. The base learning rate is set to $3 \times 10^{-2}$, with a weight decay of $5\times 10^{-2}$. We follow the Severe-Benchmark~\cite{thoker2022severe} practice and report the $30$-view top-$1$ accuracy. Other settings remain consistent with those used on UCF101.

\paragraph{Evaluation metric}
To clearly show the key factors when training self-supervised temporal tasks on top of IFMs, we evaluate the action recognition performance of model designs and training settings on UCF101, Diving48, and SSv2 datasets, following a pre-training then fine-tuning protocol. For evaluation, $\mathbf{Acc_{ft}}$ denotes the fine-tuning performance with pre-training initialization, which represents the capability of a cascade IFM-TMM model on these datasets, and $\mathbf{{\Delta}Acc}$ denotes the difference between $\mathbf{Acc_{ft}}$ and the fine-tuning performance without pre-training (\ie a randomly initialized TMM). A higher $\mathbf{{\Delta}Acc}$ indicates a more effective temporal modeling pre-training.

\begin{table}[t]
\centering
\caption{\textbf{Increasing pre-training epochs improves SSL effectiveness}, while still cannot remedy the performance drop. We attribute this to the severe degradation in from-scratch performance caused by aggressive channel compression.}
\label{tab:ft-delta-vs-epochs}
\resizebox{0.48\textwidth}{!}{\begin{tabular}{c|ccccc}
\toprule
\#Ch. dim. & 768 & 192 & 48 & 12 & 3 \\
\midrule
\multicolumn{5}{l}{\hspace{-6pt}\small{\textit{Evaluating on UCF101~\cite{soomro2012ucf101}}}} \\ 
\textcolor{gray}{from scratch} & \textcolor{gray}{86.81} & \textcolor{gray}{84.93} & \textcolor{gray}{82.99} & \textcolor{gray}{77.69} & \textcolor{gray}{62.20} \\
pt. 200ep & \cellcolor{gray!20}{-0.09} & +1.36 & +1.73 & +0.20 & +0.72  \\
pt. 800ep & +1.79 & +3.01 & +2.73 & +2.07 & +2.72 \\
\midrule
\multicolumn{5}{l}{\hspace{-6pt}\small{\textit{Evaluating on Diving48~\cite{li2018resound}}}} \\
\textcolor{gray}{from scratch} & \textcolor{gray}{69.48} & \textcolor{gray}{69.50} & \textcolor{gray}{65.96} & \textcolor{gray}{55.14} & \textcolor{gray}{30.59} \\
pt. 200ep & +0.10 & \cellcolor{gray!20}{-0.71} & \cellcolor{gray!20}{-1.41} & \cellcolor{gray!20}{-2.73} & \cellcolor{gray!20}{-1.55} \\
pt. 800ep & +2.71 & +0.79 & \cellcolor{gray!20}{-1.87} & \cellcolor{gray!20}{-0.55} & +0.26 \\
\bottomrule
\end{tabular}}
\end{table}

\begin{table}[t]
\centering
\caption{\textbf{The computational efficiency impact} of spatial and channel dimension compression on TMM. Reducing the spatial resolution (\#Sp. res.) moderately decreases computational costs, whereas reducing channel dimensions (\#Ch. dim.) does not help.}
\label{tab:computational-efficiency}
\resizebox{0.48\textwidth}{!}{\begin{tabular}{c|ccccc}
\toprule
\#Sp. res. & 14 & 9 & 5 & 3 & 1 \\
\midrule
Flops (G) & 66.18 & 27.35 & 8.44 & 3.04 & 0.34 \\
\bottomrule
\toprule
\#Ch. dim. & 768 & 192 & 48 & 12 & 3 \\
\midrule
Flops (G) & 66.18 & 63.40 & 62.71 & 62.53 & 62.49 \\
\bottomrule
\end{tabular}}
\end{table}

\begin{table}[t]
\centering
\caption{\textbf{The number of parameters and computational overload} for different settings of R3D~\cite{feichtenhofer2021large} blocks used as TMM. The first column fixes the hidden layer channel dimension to $256$, and the second column fixes the number of blocks to $1$.}
\label{tab:block-ch-vs-param-flops}
\resizebox{0.48\textwidth}{!}{\begin{tabular}{c|ccccc}
\toprule
\#Block & \cellcolor{gray!20}{1} & 2 & 4 & 6 & 8 \\
\midrule
\#Params. (M) & \cellcolor{gray!20}{1.84} & 2.95 & 5.18 & 7.41 & 9.63 \\
Flops (G) & \cellcolor{gray!20}{2.88} & 4.62 & 8.12 & 11.61 & 15.11 \\
\bottomrule
\toprule
\#Hidden ch. dim. & 64 & \cellcolor{gray!20}{256} & 512 & 1024 & 2048 \\
\midrule
\#Params. (M) & 0.94 & \cellcolor{gray!20}{1.84} & 4.07 & 12.06 & 42.21 \\
Flops (G) & 1.47 & \cellcolor{gray!20}{2.88} & 6.37 & 18.91 & 66.18 \\
\bottomrule
\end{tabular}}
\end{table}

\begin{table}[t]
    \centering
    \caption{\textbf{Applying different IFMs and their combinations to AdViSe framework.} We report the fine-tuning performance on UCF101~\cite{soomro2012ucf101} and Diving48~\cite{li2018resound}. All models are pre-trained on UCF101~\cite{soomro2012ucf101} for $800$ epochs under the setting of $8\times224^2$}\label{tab:other-ifms}
    \resizebox{0.48\textwidth}{!}{\begin{tabular}{c|ccc|cc}
    \toprule
    Exp. No. & CLIP & MAE & Dino-v2 & UCF101 & Diving48 \\
    \midrule
    $1$ & \checkmark & & & 91.9 & 74.2 \\
    $2$ & & \checkmark & & 86.8 & 75.5 \\
    $3$ & & & \checkmark & 95.3 & 82.9 \\
    $4$ & \checkmark & \checkmark & & 91.4 & 79.4 \\
    $5$ & \checkmark & & \checkmark & 94.1 & 81.3 \\
    \bottomrule
    \end{tabular}}
\end{table}

\subsection{Spatial Feature Utilization}
\label{sec:SFU}
This section examines the impact of spatial features on temporal modeling by discussing the average pooling method for spatial feature processing.

\paragraph{Finding 1} \textit{Spatial resolution is crucial for temporal modeling.} 
%

Fig.~\ref{fig:spatial-compress} shows the decline in $Acc_{ft}$ (bottom) and ${\Delta}Acc$ (top) indices on UCF101 and Diving48 datasets as spatial resolution reduces. When spatial feature compression approaches or reaches global pooling, self-supervised pre-training ceases to yield performance gains (lower ${\Delta}Acc$). Moreover, compared to UCF101, which relies more on spatial information, performance degradation on Diving48 is more pronounced.
It is surprising that temporal domain tasks are more susceptible to spatial feature compression than spatial domain tasks. We suggest this phenomenon occurs as temporal tasks rely heavily on temporal information derived from spatial features. The compression of spatial features impairs this process, leading to a larger performance drop in temporal tasks. In contrast, spatial tasks utilize information directly from spatial features, and thereby feature compression has relatively little impact on their performance.

\paragraph{Finding 2} \textit{Channel dimension is crucial for spatio-temporal representation.}

Fig.~\ref{fig:channel-compress}~(bottom) shows that the accuracy of downstream tasks on both UCF101 and Diving48 decreases with the compression of channel dimensions. Although pre-training for temporal modeling retains some validity, as shown in Fig.~\ref{fig:channel-compress}~(top).

Further experiments in Table~\ref{tab:ft-delta-vs-epochs} show that increasing pre-training epochs for temporal modeling cannot remedy the performance drop caused by channel dimension compression. Although additional pre-training enhances temporal modeling (indicated by a higher ${\Delta}Acc$), they are insufficient to counteract the performance drop in downstream tasks. Therefore, compression of channel dimensions in spatial features should be avoided.
It is noted that while compressing spatial resolution impairs the effectiveness of self-supervised learning, it results in a significant reduction in computational load, as shown in Table~\ref{tab:computational-efficiency}~(top). Therefore, in scenarios sensitive to overhead, using compressed spatial resolution remains a viable option.

\begin{table}[t]
    \centering
    \caption{\textbf{Comparison between TMM and ``average pooling" as temporal modeling operation.} We use the same pre-trained models and evaluation benchmark as in Table~\ref{tab:other-ifms}} \label{tab:necessity-of-tmm}
    \resizebox{0.48\textwidth}{!}{\begin{tabular}{cc|cc}
    \toprule
    IFM & Temporal modeling setting & UCF101 & Diving48 \\
    \midrule
    \multirow{2}{*}{CLIP} & TMM & \textbf{91.9} & \textbf{74.2} \\
     & average pooling & 88.5 & 23.1 \\
     \midrule
    \multirow{2}{*}{MAE} & TMM & \textbf{86.8} & \textbf{75.5} \\
     & average pooling & 73.0 & 14.1 \\
     \midrule
    \multirow{2}{*}{Dino-v2} & TMM & \textbf{95.3} & \textbf{82.9} \\
     & average pooling & 88.5 & 22.9 \\
    \bottomrule
    \end{tabular}}
\end{table}

\begin{table}[t]
\centering
\caption{\textbf{Comparisons of the number of parameters and memory usage}. We use ``Params." to indicate \textbf{trainable} parameters and report the GFLOPs metric of tunable module. ``Time'' indicates training time per iteration. We use a single R3D block with channel dimension $256$~/~$2048$ in TMM.}
\label{tab:memory}
\resizebox{0.48\textwidth}{!}{\begin{tabular}{c|cccc}
\toprule
Method & Params.(M) & FLops(G) & Mem.(GB) & Time(s) \\
\midrule
A-B/16-256 & 4.0 & 8.1 & 4.9 & 5.6 \\
A-L/14-256 & 4.0 & 8.1 & 6.2 & 11.8 \\ 
A-B/16-2048 & 44.3 & 66.2 & 7.9 & 17.0 \\
R3D-50~\cite{feichtenhofer2021large} & 31.8 & 125.3 & 40.0 & 19.2 \\
\bottomrule
\end{tabular}}
\end{table}

\begin{table}[t]
    \centering
    \caption{\textbf{The time consumption for inference of a video clip with input size $8\times224^2$ across different IFMs.} We use A800 for this experiment, with the data type being fp32.} \label{tab:inference-of-ifms}
    \resizebox{0.48\textwidth}{!}{\begin{tabular}{c|ccc}
    \toprule
     & CLIP-L/14 & MAE-L/16 & Dino-v2-L/14 \\
    \midrule
    Inference time(s) & 1.57 & 1.45 & 1.92 \\
    \bottomrule
    \end{tabular}}
\end{table}

\subsection{Temporal Modeling Module}
\label{sec:TMM}

This section provides key insights into temporal modeling based on the design configuration of TMM.

\paragraph{Finding 3} \textit{Increasing the depth of TMM does not always have positive impacts.}
%
As shown in Fig.~\ref{fig:incre-block}, a single R3D~\cite{feichtenhofer2021large} block achieves $87.97\%$ accuracy on UCF101, while adding more blocks the performance drops.
We speculate that once network capacity sufficiently meets the requirements for spatio-temporal information extraction, adding more layers complicates training optimization~\cite{srivastava2015training}, thereby leading to the performance drop.
As Diving48 requires more temporal understanding than UCF101, it demands greater network capacity of TMM. Consequently, increasing number of blocks (from $1$ to $4$) significantly enhances the $Acc_{ft}$ performance on Diving48.
To further investigate the impact of network width on the training effectiveness of AdViSe, we configure TMM with a single R3D block layer and gradually increase the network width to evaluate the performance on the downstream tasks UCF101 and Diving48.

\paragraph{Finding 4}\textit{Increasing the channel dimension of TMM's hidden layer enhances performance at the cost of training overhead.}

As shown in Fig.~\ref{fig:incre-channel}, increasing the channel dimension (from $64$ to $512$) of TMM's hidden layer initially leads to  downstream performance gain. However, further increasing the channel dimension results in limited gain.
The increase in channel dimensions enhances the network's capacity, and single-layer R3D~\cite{feichtenhofer2021large} blocks are easier to optimize than deeper networks, which contributes to performance gains. However, as shown in Table~\ref{tab:block-ch-vs-param-flops}, increasing the number of hidden layer channels significantly raises TMM's parameter count, which implies a higher training cost. Therefore, we recommend using a TMM with $256$ channels as the standard configuration for AdViSe while limiting the channel dimensions to no more than $2048$ in practice.

\begin{table*}[t]
\centering
\small
\caption{\textbf{Comparison of downstream task performance.} 
We compare the metrics of each method in columns according to the input resolution. IFM in AdViSe is frozen in all phases. Only TMM and classification head are tunable. * denotes the method using extra input modal (\ie ``frames diff"), and † denotes the method contains an extra tunable module without taking into account.}
\label{tab:action-recog-UCF101-h51}
\resizebox{\textwidth}{!}{\begin{tabular}{l|cccc|cc}
\toprule
Pre-trained on UCF101  & Input Frames & Backbone & \#Tr. Params. & PT. Task & UCF101 & HMDB51\\
\midrule
PRP~\cite{yao2020video}~(2020)  & $16\times112^2$ & R21D-18 & 14.4M & PRP & 72.1 & 35.0 \\
TCLR~\cite{dave2022tclr}~(2022)  & $16\times112^2$ & R21D-18 & 14.4M & Contrastive & 82.8 & 53.6 \\
TransRank\textsuperscript{*}~\cite{duan2022transrank}~(2022)  & $16\times112^2$ & R3D-18 & 20.2M & PRP & 88.5 & \textbf{63.0} \\
\textbf{AdViSe} w/ CLIP-B/16  & $8\times112^2$ & ViT-B, TMM & \textbf{4.5M} & PRP & \textbf{88.7} & 56.2 \\
\toprule
VideoMAE~\cite{tong2022videomae}~(2022)  & $16\times224^2$ & ViT-B & 87.1M & Mask Modeling & 91.3 & - \\
\textbf{AdViSe} w/ CLIP-B/16  & $8\times224^2$ & ViT-B, TMM & \textbf{4.5M} & PRP & 90.7 & 58.1 \\
\textbf{AdViSe} w/ CLIP-L/14  & $8\times224^2$ & ViT-L, TMM & \textbf{4.5M} & PRP & \textbf{91.9} & \textbf{60.7} \\
\midrule
\midrule
Pre-trained on K400  & Input Frames & Backbone & \#Tr. Params. & PT. Task & UCF101 & HMDB51\\
\midrule
TCLR~\cite{dave2022tclr}~(2022)  & $16\times112^2$ & R21D-18 & 14.4M & Contrastive & 88.2 & 60.0 \\
TransRank\textsuperscript{*}~\cite{duan2022transrank}~(2022)  & $16\times112^2$ & R21D-18 & 14.4M & PRP & 90.7 & \textbf{64.2} \\
TubeletCon~\cite{thoker2023tubelet}~(2023)  & $16\times112^2$ & R21D-18 & 14.4M & Contrastive & \textbf{91.0} & 64.1 \\
\textbf{AdViSe} w/ CLIP-B/16  & $8\times112^2$ & ViT-B, TMM & \textbf{4.5M} & PRP & 90.4 & 60.1 \\
\midrule
ASCNet~\cite{huang2021ascnet}~(2021)  & $64\times224^2$ & S3D-G & 9.6M & Contrastive & 90.8 & 60.5 \\
$\rho$BYOL~\cite{feichtenhofer2021large}~(2021)  & $16\times224^2$ & R3D-50 & 31.8M & Contrastive & 95.5 & \textbf{73.6} \\
VideoMAE~\cite{tong2022videomae}~(2022)  & $16\times224^2$ & ViT-B & 87.1M & Mask Modeling & \textbf{96.1} & 73.3 \\
HiCo++~\cite{10119224}~(2023) & $8\times224^2$ & R3D-50 & 31.8M & Contrastive & 94.9 & 71.8 \\
No-More-Shortcuts~\cite{dave2024no}~(2024) & $8\times224^2$ & ViT-L & 202.1M\textsuperscript{†} & OFL\&TSP & 94.3 & 64.3 \\
\textbf{AdViSe} w/ CLIP-B/16  & $8\times224^2$ & ViT-B, TMM & \textbf{4.5M} & PRP & 92.3 & 60.5 \\
\textbf{AdViSe} w/ CLIP-L/14  & $8\times224^2$ & ViT-L, TMM & \textbf{4.5M} & PRP & 94.1 & 63.0 \\
\bottomrule
\end{tabular}}
\end{table*}

\subsection{Discussion}
\label{sec:discussion}

We will discuss noteworthy details within the AdViSe framework, encompassing: the impact of different IFM selections, the role of PRP~\cite{yao2020video}, and the necessity of TMM.

\paragraph{Using MAE and Dino-v2 as IFMs}

MAE~\cite{HeCXLDG22} and Dino~\cite{OquabDMVSKFHMEA24} series methods have elevated the benchmark for self-supervised learning to new heights. This section discusses the performance of the AdViSe framework on downstream tasks when employing MAE and Dino-v2 as IFMs. As illustrated in Table~\ref{tab:other-ifms}, when using MAE-L/16~\cite{HeCXLDG22} as the IFM, its performance on UCF101~\cite{soomro2012ucf101} and Diving48~\cite{li2018resound} datasets is slightly inferior to that of using CLIP-L/14~\cite{radford2021learning} as the IFM~(exp.$1$ \& $2$), while Dino-v2~\cite{OquabDMVSKFHMEA24} exhibits better performance as the IFM~(exp.$1$ \& $3$). The performance disadvantage of MAE-L/16 may due to its larger patch size. Further experiments combined MAE and Dino-v2 with CLIP respectively~(exp.$4$\ \& $5$), yet failed to achieve superior performance in all settings, which might suggest that simple feature concatenation is insufficient to fully integrate the strengths of different IFMs. These experiments demonstrate that different IFMs significantly impact AdViSe's performance. The most powerful self-supervised model, such as Dino-v2~\cite{OquabDMVSKFHMEA24}, can attain performance comparable to CLIP~\cite{radford2021learning}, while further research is required to explore how to effectively integrate representation from different IFMs.

\paragraph{A Further Discussion of PRP}

In PRP task, a 3D representation model is required to predict the sampling rate of a video that has undergone temporal down-sampling~\cite{yao2020video}. Within AdViSe framework, the fundamental form of PRP is retained, where frames are first downsampled, then representations are extracted to predict the video playback speed. The distinction lies in that the weights of IFMs are frozen, so only TMM are optimized by supervision signal. This implies that the spatial feature extraction capability of IFMs is fully preserved, while TMM needs to model the temporal dynamic required for PRP task from multi-frame spatial features. We believe this will have two implications: 1. Training will focus on the integration of temporal features, thereby avoiding the hacking of ``spatial shortcuts"~\cite{dave2024no}; 2. It adds flexibility to AdViSe framework, enabling the trained TMM to become a "plug-and-play" module.

\paragraph{The Necessity of TMM}

Experiments in Sec.~\ref{sec:TMM} demonstrate that a single R3D~\cite{feichtenhofer2021large} block can achieve outstanding performance on the UCF101~\cite{soomro2012ucf101} dataset, which naturally raises a question: Is TMM necessary? Can simple temporal fusion (e.g., average pooling) achieve good performance on datasets biased towards spatial understanding~\cite{soomro2012ucf101, kuehne2011hmdb}? How does it perform on datasets emphasizing temporal understanding~\cite{li2018resound, goyal2017something}? The results presented in Table~\ref{tab:necessity-of-tmm} indicate that TMM is crucial for effectively integrating spatial representations. Under varying IFM configurations, the setups employing TMM consistently outperform those using average pooling in downstream tasks, particularly on dataset emphasizing temporal understanding.

\section{Experiment}
\label{sec:performance}

In this section, we validate the effectiveness of the proposed AdViSe approach by evaluating its training efficiency, performance on downstream tasks, and feature visualization.

\subsection{Training Efficiency}
\label{sec:effect}

We evaluate three variants of AdViSe, A-B/16-256, A-B/16-2048, and A-L/14-256, with a standard ResNet3D-50~\cite{feichtenhofer2021large} (R3D-50) in terms of training time and memory costs on A800 GPUs. 
In specific, A-B/16-256 uses CLIP-ViT-B/16~\cite{radford2021learning} as IFM and incorporates a single R3D block with channel dimension $256$ as TMM. 
We train R3D-50 using the PRP self-supervised learning method~\cite{yao2020video} with random initialization. As shown in Fig.~\ref{fig:train-eff}~(a), AdViSe variants enjoy much lower training time costs and superior performance compared to R3D-50. Specifically, compared to R3D-50, A-B/16-256 reduces training time by a factor of $3.4$ and improves downstream performance by approximately $6\%$. AdViSe models of different sizes demonstrate varying performance and overhead priorities while consistently outperforming the spatio-temporal simultaneous learning method.

We compare the AdViSe variants and R3D-50~\cite{feichtenhofer2021large} in terms of parameters and memory usage during training. As shown in Table~\ref{tab:memory}, we sample $128$ clips in a mini-batch with an input clip size of $8\times224^2$ and measure the trainable parameters, forward pass FLOPs, training time per iteration, and memory usage. We provide the time cost of different IFMs for one sample inference as a reference, as shown in Table~\ref{tab:inference-of-ifms}. We adopt the same sequential method as described in~\cite{feichtenhofer2021large} for fair comparison. Although IFMs have a large number of parameters, these frozen parameters do not increase video memory overhead significantly during training. Additionally, the trainable parameters in AdViSe (\ie TMM and MLP head) are primarily computed on the feature maps of samples, resulting in lower memory consumption due to smaller activation graphs.

\subsection{Performance on Down-stream Tasks}
\paragraph{Action Recognition on UCF101 and HMDB51}
In Table~\ref{tab:action-recog-UCF101-h51}, we present a comparative analysis of AdViSe against state-of-the-art methods for action recognition on the UCF101 and HMDB51 datasets. We use CLIP~\cite{radford2021learning} as IFM in all experiments.

Compared with PRP~\cite{yao2020video}, AdViSe achieves a notable performance enhancement. Specifically, we observe an improvement from $72.1\%$ to $88.7\%$ on UCF101, utilizing only $31.3\%$ of the trainable parameters.
%
In terms of transfer generalization, the spatio-temporal models pre-trained on K400 achieve competitive results with previous methods on UCF101 and HMDB51. Specifically, AdViSe w/ ViT-L achieves $94.1\%$ accuracy on UCF101 and $63.0\%$ on HMDB51. When compared to No-More-Shortcuts (NMS)~\cite{dave2024no}, which incorporates IFM as a training part for temporal pretext task pre-training, AdViSe achieves very close performance ($94.1\%$~vs.~$94.3\%$) with only $2.2\%$ trainable parameters. All the compared models pre-trained on K400 for 200 epochs with ViT-B (ViT-L) froze as IFM, and followed by a TMM consisting of a R3D~\cite{feichtenhofer2021large} layer with ${4.5M}$ tunable parameters ($4$ R3D blocks).

\begin{table}[t!]
    \centering
    \caption{\textbf{Fine-tuned action recognition performance on Diving48}~\cite{li2018resound} with model pre-trained on K400 dataset for 200 epochs under the setting of $16\times224^2$.
    }
    \label{tab:finetuning-d48}
        \resizebox{0.48\textwidth}{!}{\begin{tabular}{l|cc|c}
            \toprule
            Method & Backbone & \#Tr. Params. & Top1  \\
            \midrule
            VideoMAE~\cite{tong2022videomae} & ViT-B & 57.3M & 65.3 \\
            MGM~\cite{DBLP:conf/iccv/FanWLZBSML23} & ViT-B & 57.3M & 70.2 \\
            CPNet~\cite{liang2022self} & R21D-18 & 14.4M & 73.6 \\
            HiCo++~\cite{10119224} & R21D-18 & 14.4M & 81.2 \\
            \textbf{AdViSe} w/ CLIP-L/14 & ViT-L, TMM & \textbf{4.5M} & \textbf{81.6} \\
            \bottomrule
        \end{tabular}}
\end{table}

\begin{table}[t!]
    \centering
    \caption{\textbf{Fine-Tuned Action Recognition Results on SSv2}~\cite{goyal2017something} with model pre-trained using the same settings as on Diving48.}
    \label{tab:finetuning-ssv2}
        \resizebox{0.48\textwidth}{!}{\begin{tabular}{l|cc|c}
            \toprule
            Method & Backbone  & \#Tr. Params. & Top1  \\
            \midrule
            MSCL~\cite{ni2022motion} & R3D-18 & 20.2M & 50.3 \\
            HiCo++~\cite{10119224} & S3D-G & 9.6M & 54.8 \\
            RSPNet~\cite{chen2021rspnet} & S3D-G & 9.6M & 55.0 \\
            \textbf{AdViSe} w/ CLIP-L/14 & ViT-L, TMM & \textbf{4.5M} & \textbf{55.3} \\
            \bottomrule
        \end{tabular}}
\end{table}

\paragraph{Action Recognition on Diving48 and SSv2}
Tables~\ref{tab:finetuning-d48} and~\ref{tab:finetuning-ssv2} show that AdViSe is comparable to or outperforms previous methods. On Diving48, it achieves $81.6\%$ top-1 accuracy, compared to $81.2\%$ of HiCo++~\cite{10119224}. On SSv2, it achieves $55.3\%$ top-1 accuracy, compared to $55.0\%$ of RSPNet~\cite{chen2021rspnet}. We use the same pre-trained model on K400 with ViT-L as IFM.

\subsection{Visualization}

\begin{figure}[!t]
    \centering
    \includegraphics[width=1.0\linewidth]{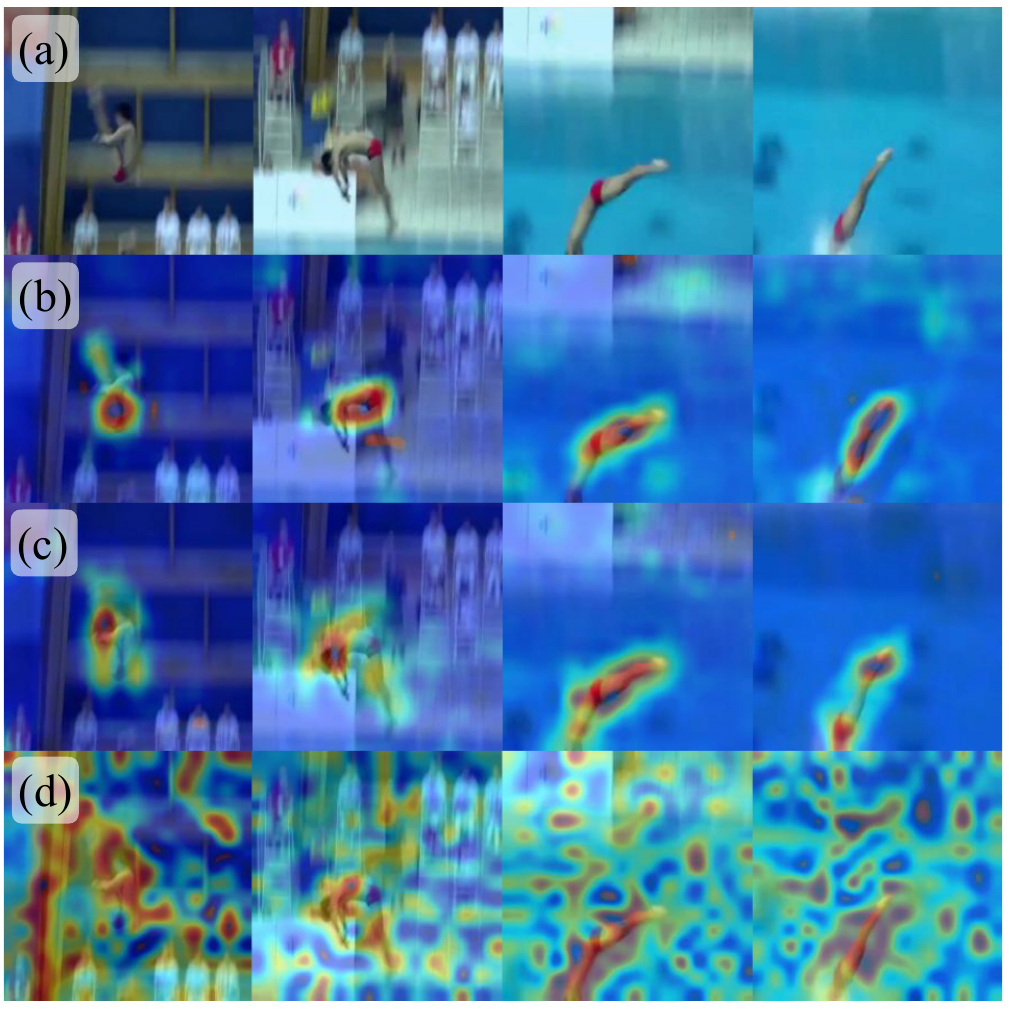}
    \caption{\textbf{Comparison of feature activation maps.} We present the feature activation maps from the last TMM layer of \textbf{(b)} AdViSe , \textbf{(c)} using TMM without PRP pre-training and \textbf{(d)} the last ViT layer of IFM~\cite{radford2021learning} on Diving48~\cite{li2018resound} dataset for \textbf{(a)} the same input clip.}
    \label{fig:visualization}
\end{figure}

We compare the feature activation maps by different methods on Diving48 dataset in Fig.~\ref{fig:visualization}, which show that AdViSe~\textbf{(b)} has a strong ability to extract dynamic information in temporal domain. We also show the feature activation maps of a variant methods~\textbf{(c)}, which uses an untrained TMM to aggregate temporal information. Additionally, we include feature activation maps from the final layer of IFM~\textbf{(d)}. As depicted in Fig.~\ref{fig:visualization}, the features by AdViSe accurately focus on the diver, paying less attention to irrelevant dynamic areas. In contrast, Scratch assigns less attention to the athlete, with some attention scattered across the scene. IFM alone is unable to effectively aggregate dynamic information, leading to the model's failure to extract dynamic areas from consecutive multi-frame images.

\section{Conclusion}
\label{sec:conclusion}

In this study, we propose the AdViSe paradigm, which directly leverages the spatial representation capability of image foundation models to train spatio-temporal representation models.
We investigate the guidelines for utilizing spatial features and designing temporal modeling modules and validate that these can significantly reduce the training overhead of video representation models using pre-trained image foundation models.
This study provides fresh insight into low-cost video self-supervised learning based on pre-trained image foundation models.

\section{Acknowledgments}
This work has been supported by the National Natural Science Foundation of China~(NSFC) under Grant No.6240071660, U24B600013.

\clearpage


\bibliographystyle{elsarticle-num} 
\bibliography{main}

\begin{thebibliography}{10}
\expandafter\ifx\csname url\endcsname\relax
  \def\url#1{\texttt{#1}}\fi
\expandafter\ifx\csname urlprefix\endcsname\relax\def\urlprefix{URL }\fi
\expandafter\ifx\csname href\endcsname\relax
  \def\href#1#2{#2} \def\path#1{#1}\fi

\bibitem{radford2021learning}
A.~Radford, J.~W. Kim, C.~Hallacy, A.~Ramesh, G.~Goh, S.~Agarwal, G.~Sastry, A.~Askell, P.~Mishkin, J.~Clark, G.~Krueger, I.~Sutskever, Learning transferable visual models from natural language supervision, in: ICML, Vol. 139, 2021, pp. 8748--8763.

\bibitem{DBLP:conf/icml/RadfordKHRGASAM21}
A.~Radford, J.~W. Kim, C.~Hallacy, A.~Ramesh, G.~Goh, S.~Agarwal, G.~Sastry, A.~Askell, P.~Mishkin, J.~Clark, G.~Krueger, I.~Sutskever, Learning transferable visual models from natural language supervision, in: ICML, Vol. 139, 2021, pp. 8748--8763.

\bibitem{schiappa2023self}
M.~C. Schiappa, Y.~S. Rawat, M.~Shah, Self-supervised learning for videos: {A} survey, {ACM} Comput. Surv. 55~(13s) (2023) 288:1--288:37.

\bibitem{feichtenhofer2021large}
C.~Feichtenhofer, H.~Fan, B.~Xiong, R.~B. Girshick, K.~He, A large-scale study on unsupervised spatiotemporal representation learning, in: IEEE CVPR, 2021, pp. 3299--3309.

\bibitem{tong2022videomae}
Z.~Tong, Y.~Song, J.~Wang, L.~Wang, Videomae: Masked autoencoders are data-efficient learners for self-supervised video pre-training, in: NeurIPS, 2022.

\bibitem{castro2022fitclip}
S.~Castro, F.~Caba, Fitclip: Refining large-scale pretrained image-text models for zero-shot video understanding tasks, in: BMVC, 2022, p. 939.

\bibitem{dave2024no}
I.~R. Dave, S.~Jenni, M.~Shah, No more shortcuts: Realizing the potential of temporal self-supervision, in: AAAI, 2024, pp. 1481--1491.

\bibitem{DBLP:conf/cvpr/ParkLS23}
J.~Park, J.~Lee, K.~Sohn, Dual-path adaptation from image to video transformers, in: IEEE CVPR, 2023, pp. 2203--2213.

\bibitem{yang2023aim}
T.~Yang, Y.~Zhu, Y.~Xie, A.~Zhang, C.~Chen, M.~Li, {AIM:} adapting image models for efficient video action recognition, in: ICLR, 2023.

\bibitem{ZhangZZSWZ24}
B.~Zhang, Y.~Zhang, J.~Zhang, Q.~Sun, R.~Wang, Q.~Zhang, Visual-guided hierarchical iterative fusion for multi-modal video action recognition, Pattern Recognit. Lett. 186 (2024) 213--220.

\bibitem{yao2020video}
Y.~Yao, C.~Liu, D.~Luo, Y.~Zhou, Q.~Ye, Video playback rate perception for self-supervised spatio-temporal representation learning, in: IEEE CVPR, 2020, pp. 6547--6556.

\bibitem{deng2009imagenet}
J.~Deng, W.~Dong, R.~Socher, L.~Li, K.~Li, L.~Fei{-}Fei, Imagenet: {A} large-scale hierarchical image database, in: IEEE CVPR, 2009, pp. 248--255.

\bibitem{soomro2012ucf101}
K.~Soomro, A.~R. Zamir, M.~Shah, {UCF101:} {A} dataset of 101 human actions classes from videos in the wild, CoRR abs/1212.0402 (2012).

\bibitem{he2020momentum}
K.~He, H.~Fan, Y.~Wu, S.~Xie, R.~B. Girshick, Momentum contrast for unsupervised visual representation learning, in: IEEE CVPR, 2020, pp. 9726--9735.

\bibitem{grill2020bootstrap}
J.~Grill, F.~Strub, F.~Altch{\'{e}}, C.~Tallec, P.~H. Richemond, E.~Buchatskaya, C.~Doersch, B.~{\'{A}}. Pires, Z.~Guo, M.~G. Azar, B.~Piot, K.~Kavukcuoglu, R.~Munos, M.~Valko, Bootstrap your own latent - {A} new approach to self-supervised learning, in: NeurIPS, 2020.

\bibitem{chen2020simple}
T.~Chen, S.~Kornblith, M.~Norouzi, G.~E. Hinton, A simple framework for contrastive learning of visual representations, in: ICML, Vol. 119 of Proceedings of Machine Learning Research, 2020, pp. 1597--1607.

\bibitem{wang2023videomae}
L.~Wang, B.~Huang, Z.~Zhao, Z.~Tong, Y.~He, Y.~Wang, Y.~Wang, Y.~Qiao, Videomae {V2:} scaling video masked autoencoders with dual masking, in: IEEE CVPR, 2023, pp. 14549--14560.

\bibitem{wang2022bevt}
R.~Wang, D.~Chen, Z.~Wu, Y.~Chen, X.~Dai, M.~Liu, Y.~Jiang, L.~Zhou, L.~Yuan, {BEVT:} {BERT} pretraining of video transformers, in: IEEE CVPR, 2022, pp. 14713--14723.

\bibitem{ZhengLLLXWZC24}
Y.~Zheng, K.~Luo, S.~Liu, Z.~Li, Y.~Xiang, L.~Wu, B.~Zeng, C.~W. Chen, {GLOCAL:} {A} self-supervised learning framework for global and local motion estimation, Pattern Recognit. Lett. 178 (2024) 91--97.

\bibitem{yue2015beyond}
J.~Y. Ng, M.~J. Hausknecht, S.~Vijayanarasimhan, O.~Vinyals, R.~Monga, G.~Toderici, Beyond short snippets: Deep networks for video classification, in: IEEE CVPR, 2015, pp. 4694--4702.

\bibitem{tran2015learning}
D.~Tran, L.~D. Bourdev, R.~Fergus, L.~Torresani, M.~Paluri, Learning spatiotemporal features with 3d convolutional networks, in: IEEE ICCV, 2015, pp. 4489--4497.

\bibitem{carreira2017quo}
J.~Carreira, A.~Zisserman, Quo vadis, action recognition? {A} new model and the kinetics dataset, in: IEEE CVPR, 2017, pp. 4724--4733.

\bibitem{qiu2017learning}
Z.~Qiu, T.~Yao, T.~Mei, Learning spatio-temporal representation with pseudo-3d residual networks, in: IEEE ICCV, 2017, pp. 5534--5542.

\bibitem{tran2018closer}
D.~Tran, H.~Wang, L.~Torresani, J.~Ray, Y.~LeCun, M.~Paluri, A closer look at spatiotemporal convolutions for action recognition, in: IEEE CVPR, 2018, pp. 6450--6459.

\bibitem{xie2018rethinking}
S.~Xie, C.~Sun, J.~Huang, Z.~Tu, K.~Murphy, Rethinking spatiotemporal feature learning: Speed-accuracy trade-offs in video classification, in: ECCV, Vol. 11219 of Lecture Notes in Computer Science, 2018, pp. 318--335.

\bibitem{szegedy2015going}
C.~Szegedy, W.~Liu, Y.~Jia, P.~Sermanet, S.~E. Reed, D.~Anguelov, D.~Erhan, V.~Vanhoucke, A.~Rabinovich, Going deeper with convolutions, in: IEEE CVPR, 2015, pp. 1--9.

\bibitem{neimark2021video}
D.~Neimark, O.~Bar, M.~Zohar, D.~Asselmann, Video transformer network, in: IEEE ICCV Workshop, 2021, pp. 3156--3165.

\bibitem{bertasius2021space}
G.~Bertasius, H.~Wang, L.~Torresani, Is space-time attention all you need for video understanding?, in: ICML, Vol. 139, 2021, pp. 813--824.

\bibitem{arnab2021vivit}
A.~Arnab, M.~Dehghani, G.~Heigold, C.~Sun, M.~Lucic, C.~Schmid, Vivit: {A} video vision transformer, in: IEEE ICCV, 2021, pp. 6816--6826.

\bibitem{liu2022video}
Z.~Liu, J.~Ning, Y.~Cao, Y.~Wei, Z.~Zhang, S.~Lin, H.~Hu, Video swin transformer, in: IEEE CVPR, 2022, pp. 3192--3201.

\bibitem{li2018resound}
Y.~Li, Y.~Li, N.~Vasconcelos, {RESOUND:} towards action recognition without representation bias, in: ECCV, Vol. 11210 of Lecture Notes in Computer Science, 2018, pp. 520--535.

\bibitem{kuehne2011hmdb}
H.~Kuehne, H.~Jhuang, E.~Garrote, T.~A. Poggio, T.~Serre, {HMDB:} {A} large video database for human motion recognition, in: IEEE ICCV, 2011, pp. 2556--2563.

\bibitem{goyal2017something}
R.~Goyal, S.~E. Kahou, V.~Michalski, J.~Materzynska, S.~Westphal, H.~Kim, V.~Haenel, I.~Fr{\"{u}}nd, P.~Yianilos, M.~Mueller{-}Freitag, F.~Hoppe, C.~Thurau, I.~Bax, R.~Memisevic, The ``something something" video database for learning and evaluating visual common sense, in: IEEE ICCV, 2017, pp. 5843--5851.

\bibitem{loshchilov2017fixing}
I.~Loshchilov, F.~Hutter, Fixing weight decay regularization in adam, CoRR abs/1711.05101 (2017).

\bibitem{thoker2022severe}
F.~M. Thoker, H.~Doughty, P.~Bagad, C.~G.~M. Snoek, How severe is benchmark-sensitivity in video self-supervised learning?, in: ECCV, Vol. 13694 of Lecture Notes in Computer Science, 2022, pp. 632--652.

\bibitem{srivastava2015training}
R.~K. Srivastava, K.~Greff, J.~Schmidhuber, Training very deep networks, in: NeurIPS, 2015, pp. 2377--2385.

\bibitem{dave2022tclr}
I.~R. Dave, R.~Gupta, M.~N. Rizve, M.~Shah, {TCLR:} temporal contrastive learning for video representation, Comput. Vis. Image Underst. 219 (2022) 103406.

\bibitem{duan2022transrank}
H.~Duan, N.~Zhao, K.~Chen, D.~Lin, Transrank: Self-supervised video representation learning via ranking-based transformation recognition, in: IEEE CVPR, 2022, pp. 2990--3000.

\bibitem{thoker2023tubelet}
F.~M. Thoker, H.~Doughty, C.~G.~M. Snoek, Tubelet-contrastive self-supervision for video-efficient generalization, in: IEEE ICCV, 2023, pp. 13766--13777.

\bibitem{huang2021ascnet}
D.~Huang, W.~Wu, W.~Hu, X.~Liu, D.~He, Z.~Wu, X.~Wu, M.~Tan, E.~Ding, Ascnet: Self-supervised video representation learning with appearance-speed consistency, in: IEEE ICCV, 2021, pp. 8076--8085.

\bibitem{10119224}
Z.~Qing, S.~Zhang, Z.~Huang, Y.~Xu, X.~Wang, C.~Gao, R.~Jin, N.~Sang, Self-supervised learning from untrimmed videos via hierarchical consistency, IEEE Trans. Pattern Anal. Mach. Intell. 45~(10) (2023) 12408--12426.

\bibitem{HeCXLDG22}
K.~He, X.~Chen, S.~Xie, Y.~Li, P.~Doll{\'{a}}r, R.~B. Girshick, Masked autoencoders are scalable vision learners, in: IEEE CVPR, 2022, pp. 15979--15988.

\bibitem{OquabDMVSKFHMEA24}
M.~Oquab, T.~Darcet, T.~Moutakanni, H.~V. Vo, M.~Szafraniec, V.~Khalidov, P.~Fernandez, D.~Haziza, F.~Massa, A.~El{-}Nouby, M.~Assran, N.~Ballas, W.~Galuba, R.~Howes, P.~Huang, S.~Li, I.~Misra, M.~Rabbat, V.~Sharma, G.~Synnaeve, H.~Xu, H.~J{\'{e}}gou, J.~Mairal, P.~Labatut, A.~Joulin, P.~Bojanowski, Dinov2: Learning robust visual features without supervision, Trans. Mach. Learn. Res. 2024 (2024).

\bibitem{DBLP:conf/iccv/FanWLZBSML23}
D.~Fan, J.~Wang, S.~Liao, Y.~Zhu, V.~Bhat, H.~J. Santos{-}Villalobos, R.~MV, X.~Li, Motion-guided masking for spatiotemporal representation learning, in: IEEE ICCV, 2023, pp. 5596--5606.

\bibitem{liang2022self}
H.~Liang, N.~Quader, Z.~Chi, L.~Chen, P.~Dai, J.~Lu, Y.~Wang, Self-supervised spatiotemporal representation learning by exploiting video continuity, in: AAAI, 2022, pp. 1564--1573.

\bibitem{ni2022motion}
J.~Ni, N.~Zhou, J.~Qin, Q.~Wu, J.~Liu, B.~Li, D.~Huang, Motion sensitive contrastive learning for self-supervised video representation, in: ECCV, Vol. 13695 of Lecture Notes in Computer Science, 2022, pp. 457--474.

\bibitem{chen2021rspnet}
P.~Chen, D.~Huang, D.~He, X.~Long, R.~Zeng, S.~Wen, M.~Tan, C.~Gan, Rspnet: Relative speed perception for unsupervised video representation learning, in: AAAI, 2021, pp. 1045--1053.

\end{thebibliography}






\end{document}